\documentclass{article}

\PassOptionsToPackage{numbers, compress}{natbib}

\usepackage[dblblindworkshop, final]{neurips_2025}

\workshoptitle{MATH-AI}

\usepackage{graphicx}

\usepackage[utf8]{inputenc} 
\usepackage[T1]{fontenc}    
\usepackage{hyperref}       
\usepackage{url}            
\usepackage{booktabs}       
\usepackage{amsfonts}       
\usepackage{nicefrac}       
\usepackage{microtype}      
\usepackage{xcolor}         

\usepackage{amsmath}
\usepackage{enumitem}
\usepackage{booktabs}
\usepackage{soul} 
\usepackage{makecell}
\usepackage{tabularx} 
\usepackage{hyperref}

\definecolor{raise}{RGB}{224, 241, 225}
\definecolor{raise2}{RGB}{224, 241, 225}
\definecolor{rag}{RGB}{230, 230, 230}
\sethlcolor{raise2} 
\definecolor{raise_jh}{RGB}{0,137,9}
\definecolor{NavyBlue}{RGB}{8,111,189}

\usepackage[ruled,vlined]{algorithm2e}

\title{RAISE: Enhancing Scientific Reasoning in LLMs via \\Step-by-Step Retrieval}

\author{%
  \textbf{Minhae Oh\textsuperscript{1}},
 \textbf{Jeonghye Kim\textsuperscript{2}},
 \textbf{Nakyung Lee\textsuperscript{1}},
 \textbf{Donggeon Seo\textsuperscript{3}},
 \\
 \textbf{Taeuk Kim\textsuperscript{3,*}},
 \textbf{Jungwoo Lee\textsuperscript{1,*}}
\\
\\
 \textsuperscript{1}Seoul National University,
 \textsuperscript{2}KAIST,
 \textsuperscript{3}Hanyang University
\\} 

\begin{document}

\maketitle

\begin{abstract}
  Scientific reasoning requires not only long-chain reasoning processes, but also knowledge of domain-specific terminologies and adaptation to updated findings. To deal with these challenges for scientific reasoning, we introduce \textbf{RAISE}, 
a step-by-step retrieval-augmented framework which retrieves logically relevant documents from in-the-wild corpus. RAISE is divided into three steps: problem decomposition, logical query generation, and logical retrieval. We observe that RAISE consistently outperforms other baselines on scientific reasoning benchmarks. We analyze that unlike other baselines, RAISE retrieves documents that are not only similar in terms of the domain knowledge, but also documents logically more relevant. Codes are available at \href{https://github.com/minhaeoh/RAISE}{https://github.com/minhaeoh/RAISE}.
\end{abstract}

\section{Introduction}
Large language models (LLMs) have shown strong potential for scientific reasoning, which demands advanced reasoning skills, domain-specific terminology, and up-to-date knowledge \citep{zhang2024scientificlargelanguagemodels,zhang2024sciglm,prabhakar2025omnisciencedomainspecializedllmscientific,rueda2025understandingllmscientificreasoning}. Two common strategies are step-wise reasoning, which solves complex problems through structured intermediate steps \citep{cot,ssc-cot,impact-step-length,step-dpo,tot}, and retrieval-augmented generation (RAG), which mitigates hallucinations by providing external evidence \citep{rag,hallulens,zhong2025benchmarkingretrievalaugmentedgenerationchemistry,xiong2024benchmarkingretrievalaugmentedgenerationmedicine}. 
Recent work combines them, but often targets simpler multi-hop QA or assumes curated, task-specific corpora \citep{jeong2024adaptive,zhang2025credible,guan2025deepragthinkingretrievalstep,jin2025search,wang2025chain}, unlike open-domain sources such as Wikipedia.
Solving challenging scientific reasoning tasks, such as graduate-level biology or chemistry, using an in-the-wild corpus is difficult since merely retrieving superficial knowledge is insufficient. 
Instead, the retrieved information should contain relevant logical connections needed to solve the problem \citep{rueda2025understandingllmscientificreasoning}. 
Moreover, the knowledge required for each intermediate step can vary significantly even within the same problem. 
Without considering the evolving information needed for each reasoning process, RAG might even deteriorate the downstream task performance. 
The question of \textit{what to search for} and \textit{how to retrieve} the appropriate external knowledge for each step when solving scientific reasoning tasks is underexplored. 

\begin{figure*}[t!]
    \centering
    \includegraphics[width=\textwidth]{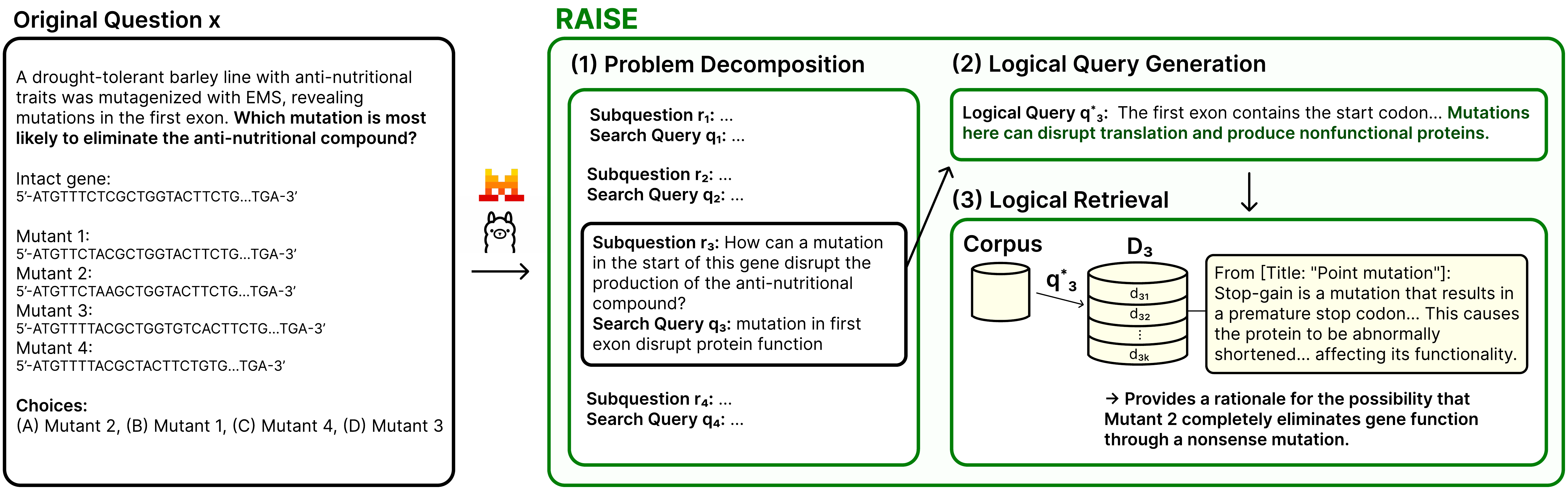}
    \caption{\textbf{Overview of RAISE.} RAISE is divided into three steps: (1) Problem Decomposition, (2) Logical Query Generation, and (3) Logical Retrieval. }
    \label{fig:raise}
\end{figure*}


To address these challenges, we introduce \textbf{RAISE} (Step-by-Step \textbf{R}etrieval-\textbf{A}ugmented \textbf{I}nference for \textbf{S}cientific r\textbf{E}asoning), a retrieval-augmented framework tailored for step-wise scientific reasoning. 
Our framework consists of three stages: (1) problem decomposition, where LLMs break down the original question into subquestions along with search queries; (2) logical query generation, which reformulates each search query into a logic-enriched query that captures the reasoning needed to solve the subquestion; and (3) logical retrieval, which retrieves step-specific documents from an open-domain corpus, ensuring the retrieved information is logically relevant rather than superficially domain similar.
Instead of assuming task-relevant or well-curated retrieval source, such as question-answer pool of relevant domains, we retrieve from in-the-wild source such as Wikipedia, which enables applying to challenging real-world scenarios. 
Evaluated on GPQA, SuperGPQA, and MMLU, RAISE consistently outperforms baselines using either RAG or problem decomposition alone, demonstrating its ability to retrieve step-specific, logically relevant information essential for solving complex scientific reasoning tasks. While this work focuses on scientific reasoning benchmarks, similar challenges arise in mathematical reasoning, which also requires precise, multi-step logical inference.

\section{Preliminary} \label{appendix:preliminary}
\paragraph{Step-by-Step Reasoning in LLMs.}
LLMs are capable of performing multi-step reasoning over complex input queries by internally chaining intermediate inferences. This step-by-step reasoning process involves decomposing a question into sub-problems, maintaining coherence across steps, and generating a final answer. Formally, given a query \( x \), the model implicitly constructs a latent reasoning trajectory \( \{r_t\}_{t=1}^T \), and generates the answer \( y \) conditioned on this chain:
\[
p(y \mid x) = \sum_{r_1, \dots, r_T} p(y \mid r_{1:T}, x) \cdot \prod_{t=1}^{T} p(r_t \mid r_{<t}, x).
\]
However, standard LLMs rely solely on their parametric knowledge, which limits performance in scenarios requiring up-to-date or external information.  

\paragraph{RAG for Single-Step Reasoning.} 
We address the task of generating a response \( y \) given an input \( x \), enhanced by retrieval from an external corpus \( \mathcal{D} \). RAG combines a retriever and a generator to condition the output on both the input and relevant documents.

A standard language model defines:
\[
p(y \mid x) = \prod_{t=1}^{T} p(y_t \mid y_{<t}, x).
\]
In RAG, generation is conditioned on retrieved documents \( \{d_j\}_{j=1}^k \), typically approximated as:
\[
p(y \mid x) \approx \sum_{j=1}^k p(y \mid x, d_j) \cdot p(d_j \mid x).
\]
The retriever encodes queries and documents via \( f_q(x) \) and \( f_d(d) \), scoring relevance by:
\[
\text{sim}(x, d) = f_q(x)^\top f_d(d).
\]
Top-\( k \) documents are retrieved, and a generator (e.g., BART \citep{bart}, T5 \citep{t5}) produces \( y \) based on both \( x \) and \( d_i \).

\paragraph{Retrieval in In-the-Wild Settings.} 
We use the term “in-the-wild” to refer to open-domain corpora like Wikipedia that are not tailored for specific tasks or domains. Unlike curated corpora, they require retrieving logically relevant evidence from a large, diverse, and often tangential pool of content, making retrieval and reasoning more challenging.

\section{RAISE}

We propose \textbf{RAISE} (Step-by-Step \textbf{R}etrieval-\textbf{A}ugmented \textbf{I}nference for \textbf{S}cientific r\textbf{E}asoning), a retrieval-augmented generation framework for scientific reasoning designed to support multi-step reasoning through fine-grained, step-aware retrieval.  The method consists of three main stages: (1) Problem Decomposition, (2) Logical Query Generation, (3) Logical Retreival. The overview of RAISE is provided in Figure \ref{fig:raise} and Algorithm \ref{algo:raise} in Appendix \ref{appendix:algorithm}. 

\paragraph{Problem Decomposition.} 
RAISE decomposes the problem $x$ into subquestions $r_1, \dots, r_n$ with corresponding search queries $q_1, \dots, q_n$, forming a structured sequence for step-wise retrieval, unlike conventional single-query approaches.
These queries are not used directly for retrieval but rather serve as an initial query for the next stage. As a result, this stage outputs subquestion-query pairs $\{(r_i, q_i)\}_{i=1}^{n}$, forming the basis for step-wise retrieval and generation.



\paragraph{Logical Query Generation.} 
In the second stage, each initial search query $q_i$ and its corresponding subquestion $r_i$ are jointly used to generate a logically enriched \textbf{logical query} $q_i^*$. 
Since initial queries $q_i$ lacks reasoning context and subquestions $r_i$ alone can be noisy or overly specific, neither $q_i$ nor $r_i$ alone is sufficient for effective retrieval. By combining both, we generate logical queries that better capture the reasoning intent and retrieve logically relevant knowledge for solving each step. The model is prompted with both $q_i$ and $r_i$, along with a reformulation prompt $p_2$. 
Even if the reformulated query $q_i^*$ contains factual inaccuracies, it tends to retrieve passages from a corpus $\mathcal{C}$ that are logically relevant and supportive of the reasoning required for solving the original problem. Figure \ref{fig:query_ex} in Appendix \ref{appendix:qualitative_evaluation} presents example queries generated by RAISE, Step-Back+RAG, and HyDE, illustrating RAISE's ability to generate logical queries that are well-aligned with the reasoning intent.

\paragraph{Logical Retrieval.} 
External knowledge $D_i$ is retrieved for each subquestion $r_i$ from in-the-wild corpus $\mathcal{C}$ (e.g., Wikipedia) and used to generate the subanswer $a_i$. 
We retrieve background knowledge for each subquestion using a similarity threshold $T$ to filter irrelevant documents.
After retrieval, for each subquestion $r_i$, the model predicts its solution $a_i$ using $D_i$, the original question $x$, and the previous steps.
Finally, all subanswers are combined to generate the final answer $y$.

\newcolumntype{P}[1]{>{\centering\arraybackslash}p{#1}}
\newcolumntype{Y}{>{\centering\arraybackslash}X}

\begin{table}[t!]
    \centering
    \small
    \renewcommand{\arraystretch}{1.1}
    \setlength{\tabcolsep}{3pt}

    \begin{tabularx}{\textwidth}{l c c c c c c c}
        \toprule
        & \multicolumn{1}{c}{\textbf{GPQA}} 
        & \multicolumn{3}{c}{\textbf{SuperGPQA}}
        & \multicolumn{3}{c}{\textbf{MMLU}} \\
        \cmidrule(lr){2-2} \cmidrule(lr){3-5} \cmidrule(lr){6-8}
        & Overall & \makecell{science-\\hard} & \makecell{science-\\middle}  & \makecell{engineering-\\hard} & \makecell{(Pro)\\Chemistry} & \makecell{(Pro)\\Biology}  & \makecell{(STEM)\\College Chemistry} \\
        \midrule
        \multicolumn{8}{l}{\textbf{Direct}} \\
        CoT & 42.42 & 4.52 & 15.08 & 6.53 & \underline{25.44} & 51.88 & \underline{49.50}\\
        \midrule
        \multicolumn{8}{l}{\textbf{Direct+RAG}} \\
        CoT+RAG & 45.96 & \underline{7.54} & 12.56 & 7.54 & 25.18 & 54.39 & 43.00 \\

        \midrule
        \multicolumn{8}{l}{\textbf{Decomposed}} \\
        Least-to-Most & 44.95 & 6.03 & 14.57 & \underline{10.05} & 24.56 & 53.97 & 45.40 \\
        Step-Back & 44.44 & 5.03 & 15.08 & 6.03 & 22.70 & 56.49 & 43.00 \\
        \midrule
        \multicolumn{8}{l}{\textbf{Decomposed+RAG}} \\
        Least-to-Most+RAG & 45.95 & 6.03 & 14.57 & 8.04 & 22.97 & \underline{58.02} & 46.00 \\
        Step-Back+RAG & 43.43 & 5.53 & \underline{15.58} & 9.05 & 23.06 & 56.34 & 43.00 \\
        HyDE & \underline{46.46} & \underline{7.54} & 13.07 & 7.04 & 22.97 & 57.88 & 49.00 \\
        \midrule

        \multicolumn{8}{l}{\textbf{Ours}} \\
        RAISE 
        & \makecell{\textbf{51.01} \\{(+9.8\%)}} 
        & \makecell{\textbf{10.05} \\ {(+33.3\%)}}
        & \makecell{\textbf{19.60} \\ {(+25.8\%)}} 
        & \makecell{\textbf{10.55} \\ {(+5.0\%)}} 
        & \makecell{\textbf{28.36} \\ {(+11.5\%)}}
        & \makecell{\textbf{59.27} \\ {(+2.2\%)}} 
        & \makecell{\textbf{51.00} \\ {(+3.0\%)}}\\
        \bottomrule
    \end{tabularx}
    \vspace{0.1cm}
    \caption{Comparison of various reasoning strategies across GPQA, SuperGPQA, and MMLU. The underscore marks the best baseline, boldface the best overall, and parentheses show RAISE's gain over the top baseline. RAISE consistently outperforms other approaches for scientific reasoning benchmarks.}
    \label{tab:main_results}
\end{table}

\section{Experiment}

\subsection{Experimental Setup}
\paragraph{Datasets.} We evaluate on three scientific benchmarks: \textbf{GPQA}, \textbf{SuperGPQA}, \textbf{MMLU}, which cover graduate-level STEM and professional science tasks that require multi-step scientific reasoning.



\paragraph{Retriever and Language Models.} We adopt Dense Passage Retrieval (DPR) \citep{dpr} trained on the Natural Questions (NQ) dataset \citep{nq} as our retriever. For GPQA, our primary benchmark, we use  Mistral Small 3.1-Instruct-2503 \citep{mistral2025small31} (24B), while for SuperGPQA and MMLU we use the lighter LLaMA 3.1-8B model \citep{llama3} due to computational limits. 


\paragraph{Baselines.} 
To assess the importance of multi-step reasoning and step-aware retrieval, we conduct experiments with four groups of baselines: \textbf{Direct Reasoning}(CoT \citep{cot}), \textbf{Direct Reasoning with RAG}(CoT+RAG \citep{rag}), \textbf{Decomposed Reasoning}(Least-to-Most \citep{leasttomost}, Step-Back \citep{step-back}), and \textbf{Decomposed Reasoning with RAG}. The last group retrieves evidence for each subquestion and solves them step-by-step. 
This group includes Least-to-Most+RAG \citep{rag}, Step-Back+RAG, and HyDE \citep{hyde}, with the latter two improving retrieval relevance through query reformulation, making them strong baselines.
Further details about datasets, retriever, model settings, and baselines are provided in Appendix \ref{appendix:experiment_details}.

\subsection{Main Results}

As shown in Table \ref{tab:main_results}, our proposed method, RAISE, consistently outperforms all baseline reasoning strategies across three benchmark datasets of varying difficulty: GPQA, SuperGPQA, and MMLU, achieving an average performance improvement of 13\% over the best baseline scores. Unlike other baselines whose performance varies depending on the dataset's difficulty or type, RAISE consistently demonstrates robust performance and outperforms them across different domains, types, and levels of difficulty. 
Furthermore, we confirm that these improvements hold across models of different scales, including smaller LLaMA-8B and GPT-4o mini, as shown in Appendix \ref{appendix:gpqa_various_llm}, demonstrating that RAISE’s effectiveness is not tied to a specific LLM architecture or size.

To assess the effectiveness of our logical query generation, we compare RAISE with three RAG-based decomposed reasoning baselines that differ in how they construct retrieval queries. Least-to-Most+RAG uses the subquestion itself as the query, Step-Back+RAG abstracts a general principle from the subquestion, and HyDE generates a hypothetical answer to use as the retrieval query. RAISE consistently outperforms all baselines across benchmarks, demonstrating the advantage of generating logically grounded queries that better align with the reasoning required to solve each subquestion. These results confirm that RAISE's queries go beyond retrieving documents that are merely domain-relevant or superficially similar, enabling access to knowledge that is logically aligned with the problem-solving process. Qualitative examples further support this finding, as shown in Appendix \ref{appendix:qualitative_evaluation}, where RAISE retrieves passages containing essential scientific mechanisms while conventional RAG often returns vague or unrelated content.


Unlike RAISE, decomposed reasoning methods do not always yield better performance, particularly for smaller open-source LLMs that lack sufficient background knowledge \citep{hosseini2024not, xu2025phi4minireasoningexploringlimitssmall}. 
While decomposition can help structure reasoning, without access to relevant external knowledge, smaller models may produce hallucinations or unsupported intermediate steps, sometimes leading to worse performance than direct reasoning. 
Moreover, even when retrieval is added, naive RAG can introduce additional noise. 
In such cases, the retriever may surface superficially related or distracting content rather than the core principles needed for reasoning, which can ultimately harm performance. 
This highlights the importance of retrieving logically relevant knowledge rather than merely domain-related content, a challenge that RAISE directly addresses.

\section{Analysis of RAISE}

\begin{figure}[ht!]
\centering
\includegraphics[width=\columnwidth]{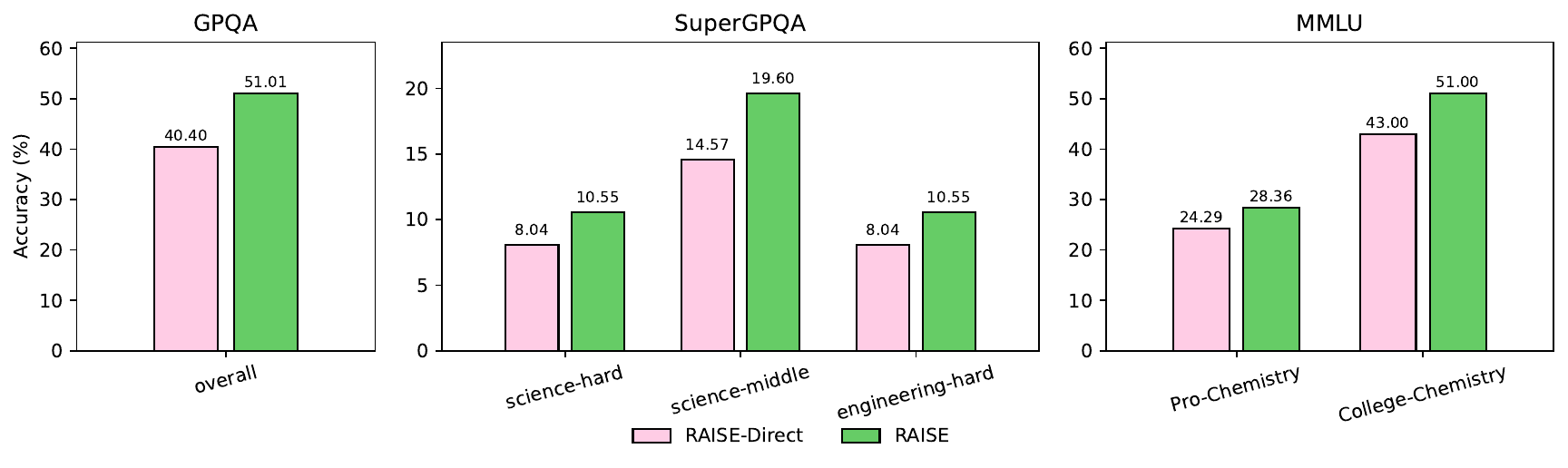}
\caption{Performance comparison between RAISE-Direct and RAISE across datasets.}
\label{fig:PD_split}
\end{figure}

To further assess the importance of problem decomposition, we also evaluate a variant of our method that omits this step and directly performs logical query generation and retrieval without breaking the problem into subquestions, as shown in Figure \ref{fig:PD_split}. This version, referred to as RAISE-Direct, showed lower performance compared to the full version of RAISE. These results indicate that problem decomposition plays a critical role in guiding the retrieval process and structuring the reasoning pathway. This suggests that for complex reasoning problems, decomposing the question and retrieving logical knowledge tailored to each subquestion is more effective than retrieving once based on the original question alone. This is likely because different reasoning steps often require distinct pieces of information that may not be jointly retrievable from a single query. 

We also analyze the quality of the retrieved documents. Using both an LLM-as-a-judge and a small-scale human evaluation, we find that RAISE consistently retrieves fewer irrelevant or superficial documents and more passages that directly support reasoning (Appendix \ref{appendix:logical_evaluation}). These results confirm that RAISE’s gains stem from retrieving logically aligned knowledge rather than merely domain-related content. 

\section{Conclusion}

We introduce RAISE, a step-by-step retrieval framework for scientific reasoning. We first decompose the problem into multiple subquestions and search queries, and then generate logical queries and  retrieve logically relevant documents from in-the-wild corpus. We demonstrate the effectiveness of RAISE on three scientific reasoning benchmarks by comparing with various baselines. Our analysis shows that RAISE retrieves documents that are not only relevant in terms of the domain (e.g. definition of specificalized terminology) but also logically relevant documents for each subquestion, assisting the step-by-step reasoning process required for scientific reasoning. 
Although our experiments focus on scientific reasoning, the stepwise logical retrieval in RAISE is broadly applicable to other domains such as mathematical problem solving, which also demands precise multi-step inference.

\section*{Acknowledgments}

This work is in part supported by the National Research Foundation of Korea (NRF, RS-2024-00451435(20\%), RS-2024-00413957(20\%)), Institute of Information \& communications Technology Planning \& Evaluation (IITP, RS-2021-II212068(10\%), RS-2025-02305453(15\%), RS-2025-02273157(15\%), RS-2025-25442149(10\%) RS-2021-II211343(10\%)) grant funded by the Ministry of Science and ICT (MSIT), Institute of New Media and Communications(INMAC), and the BK21 FOUR program of the Education, Artificial Intelligence Graduate School Program (Seoul National University), and Research Program for Future ICT Pioneers, Seoul National University in 2025.

\newpage{}
\bibliographystyle{plainnat}
\bibliography{custom}

\newpage
\appendix

\section{RAISE Algorithm} \label{appendix:algorithm}

\begin{algorithm}[ht]
\caption{RAISE Inference Procedure}
\label{alg:RAISE}
\SetAlgoLined
\DontPrintSemicolon
\KwIn{Original question $x$, prompts $\mathcal{P} = \{p_1, p_2, p_3, p_4\}$, corpus $\mathcal{C}$}
\KwOut{Final answer $y$}

\textbf{Step 1: Problem Decomposition} \;
Generate subquestions and initial queries:\;
\quad $\{(r_i, q_i)\}_{i=1}^n \sim P_\theta(\cdot \mid x, p_1)$

\For{$i = 1$ \KwTo $n$}{
  \textbf{Step 2: Logical Query Generation} \;
  Reformulate initial query:\;
  \quad $q_i^* \sim P_\theta(\cdot \mid r_i, q_i, p_2)$

  \textbf{Step 3: Knowledge Retrieval} \;
  Retrieve top-$k$ documents:\;
  \quad $D_i = \mathcal{R}(q_i^*, \mathcal{C}, k)$

  \textbf{Step 4: Subquestion Answering} \;
  \eIf{$i = 1$}{
    \quad $a_i \sim P_\theta(\cdot \mid x, r_1, D_1, p_3)$
  }{
    \quad $a_i \sim P_\theta(\cdot \mid x, \{(r_j, a_j)\}_{j=1}^{i-1}, r_i, D_i, p_3)$
  }
}

\textbf{Step 5: Final Answer Composition} \;
Generate final answer using all subanswers:\;
\quad $y \sim P_\theta(\cdot \mid x, \{(r_i, a_i)\}_{i=1}^n, p_4)$
\label{algo:raise}

\end{algorithm}

\section{Experiment Details}
\label{appendix:experiment_details}

\subsection{Dataset Details} \label{appendix:dataset_details}

\paragraph{GPQA} \citep{gpqa}
This dataset consists of physics, biology, and chemistry questions written by domain experts. We use GPQA diamond subset, which consist of 198 high-quality questions selected based on human performance. Specifically, this subset includes questions that both experts answer correctly while the majority of non-experts fail to solve. Each question typically demands multi-step reasoning, precise formula manipulation, and access to external scientific facts (e.g., physical constants, definitions). Due to its alignment with our target setting, GPQA serves as the primary evaluation benchmark throughout our experiments.

For GPQA, the original dataset does not include standardized multiple-choice labeled as (A), (B), (C) and (D). To ensure consistency during evaluation, we proprocessed each question by randomly shuffling the correct answer along with the three distractors, and assigning them uniformly to choice labels (A) through (D).

\paragraph{SuperGPQA} \citep{supergpqa} 
SuperGPQA is a large-scale benchmark designed to evaluate graduate-level reasoning across 13 disciplines, 72 fields, and 285 graduate-level disciplines. In alignment with the scientific reasoning focus of our work, we select  science and engineering domains for evaluation. Each domain is further divided by three difficulty levels(easy, medium, and hard). To reduce computational overhead while maintaining consistency, we randomly sample 199 questions per subset using a fixed seed (42). Specifically, our experiments include 199 examples each from science-hard, science-middle, and engineering-hard subsets.

\paragraph{MMLU} \citep{hendrycks2021measuringmassivemultitasklanguage, wang2024mmluprorobustchallengingmultitask} 
The MMLU benchmark covers a wide range of subjects across multiple domains. For out experiments, we focus on \textbf{STEM} and \textbf{Pro}fessional categories. The STEM contains university-level science and engineering subjects such as college mathematics and computer science, while the Professional category covers specialized fields that typically require professional training or advanced education, including law, medicine, and chemistry. We specifically select three subsets: college chemistry from MMLU-STEM and professional chemistry and biology from MMLU-Pro. These subsets are chosen to evaluate our method’s ability to perform scientific reasoning in both academic and professional contexts involving complex domain knowledge.

\subsection{Baseline Details} \label{appendix:baseline}
\textbf{CoT} \citep{cot, kojima2023largelanguagemodelszeroshot} We apply Chain-of-Thought prompting for direct reasoning, where the model is encouraged to explicitly generate intermediate reasoning steps through prompting (Think step by step).

\paragraph{CoT+RAG} \citep{rag} We implement CoT+RAG by combining Chain-of-Thought prompting with retrieval, where the model is prompted to solve the problem step-by-step while also leveraging external knowledge. Specifically, we provide the model with a CoT-style prompt encouraging step-by-step reasoning, alongside the original question and documents retrieved using the original question as the search query. 

\paragraph{Least-to-Most} \citep{leasttomost} Least-to-Most is a decomposed reasoning strategy that breaks down a complex problem into a sequence of simpler subquestions, which are then solved sequentially without retrieval augmentation. This subquestion decomposition pipeline serves as the foundational structure for other decomposed reasoning methods as well.

\paragraph{Step-Back} (Decomposed reasoning) \citep{step-back} We implement Step-Back for decomposed reasoning by applying the Step-Back prompting method to each subquestion in a decomposed reasoning framework. While the original Step-Back paper does not cover the application of this method to decomposed subquestions, we extend it for a fair comparison with our approach. Specifically, after decomposing the original question into subquestions, we use the Step-Back prompting strategy to extract a high-level principle for each subquestion, and then provide the subquestion along with its corresponding principle to guide the model’s reasoning.

\paragraph{Least-to-Most+RAG} (Decomposed reasoning with RAG) \citep{hoprag} We implement RAG by first decomposing the original problem into subquestions and then retrieving documents using each subquestion as a query. The retrieved documents are provided to the model along with the corresponding subquestion to support its reasoning.

\paragraph{Step-Back+RAG} (Decomposed reasoning with RAG) \citep{step-back} We extend the Step-Back prompting strategy to a retrieval-augmented setting for fair comparison with our method. After decomposing the original question into subquestions, we generate a principle abstraction for each subquestion using Step-Back prompting, and use it as a query to retrieve evidence. The retrieved documents are then provided alongside the original subquestion to guide the model’s reasoning.

\paragraph{HyDE} (Decomposed reasoning with RAG) \citep{hyde} We apply the HyDE approach to each subquestion in a decomposed reasoning framework. For each subquestion, the model first generates a hypothetical answer, which is then used as a query to retrieve supporting documents. The retrieved evidence, together with the subquestion, is provided to the model to support step-by-step reasoning.

\subsection{Retriever Configuration} \label{appendix:Rag_config}
We use the pre-trained DPR encoder from the 'facebook/dpr-question\_encoder-single-nq-base' model \citep{dpr}, which is a BERT-based encoder trained for open-domain question answering. This encoder is trained on the Natural Question (NQ) dataset \citep{nq} and is designed to map questions into 768-dimensional dense vector representations for retrieval. 

For the retrieval corpus, we use the preprocessed Wikipedia passages provided by 'facebook/wiki\_dpr' \citep{dpr}, a corpus widely used to evaluate DPR-based retrieval models. This corpus is constructed from the December 20, 2018 Wikipedia dump, where each article is split into multiple, disjoint text blocks of 100 words, resulting in approximately 21 million passages. Each passage is accompanied by the title of the wikipedia page it comes from along with DPR embedding.

To enable efficient retrieval over the passage embeddings, we use an exact FAISS index. FAISS (Facebook AI Similarity Search) \citep{johnson2019billion,douze2024faiss} is widely used library for fast similarity search over dense vectors. 

Throughout all experiment, we retrieve top-10 documents per query. To reduce the impact of potentially irrelevant documents by DPR, we apply a similarity threshold $T$ in RAISE. Specifically, we discard any retrieved passage whose DPR similarity score falls below $T$. DPR similarity is computed as the inner product between L2-normalized query and passage embeddings. Higher scores indicate greater semantic similarity, with values closer to 1 representing stronger alignment between the query and passage. We set $T=0.84$ for GPQA, SuperGPQA, and MMLU-Pro, which are composed of more challenging reasoning problems. For MMLU-STEM (college chemistry), we use a slightly lower threshold of $T=0.80$, considering that the questions are generally simpler than those in other datasets.

\section{Additional Results}

\subsection{Applying RAISE to various LLMs.} \label{appendix:gpqa_various_llm}

To assess the generalizability of RAISE across different LLM scales, we evaluate its performance on GPQA using LLaMA 3.1-8B \citep{llama3} and GPT-4o mini \citep{gpt-4o}, in addition to Mistral (used in our main experiments). As shown in Table \ref{tab:vertical-llm-eval}, RAISE demonstrates consistent improvements over other baselines, exhibiting a similar trend to our main results with Mistral-24B. This shows that the effect of RAISE is not limited to a specific type of LLM, but can be applied to various LLMs with different scales.

\begin{table}[h!]
    \centering
    \renewcommand{\arraystretch}{1.2}
    \setlength{\tabcolsep}{5pt}
    
    \begin{tabular}{p{3cm}P{1.3cm}P{1.3cm}P{1.3cm}}
        \toprule
         & \textbf{LLaMA} & \textbf{GPT} & \textbf{Mistral} \\
        \midrule
        \multicolumn{4}{l}{\textbf{Direct}} \\
        CoT & 22.22 & 40.91 &42.42 \\
        \midrule
        \multicolumn{4}{l}{\textbf{Direct+RAG}} \\
        CoT+RAG & 23.23 & 40.40 & 45.96\\
        \midrule
        \multicolumn{4}{l}{\textbf{Decomposed}} \\
        Least-to-Most & 26.26 & \underline{45.45} &44.95 \\
        Step-Back & \underline{28.28} & 42.42 &44.44 \\
        \midrule
        \multicolumn{4}{l}{\textbf{Decomposed+RAG}} \\
        Least-to-Most+RAG & 24.24 & 42.93 & 45.95 \\
        Step-Back+RAG & 21.72 & 42.42 &43.43\\
        HyDE & 25.75 & 38.89 &\underline{46.46} \\
        \midrule
        \multicolumn{4}{l}{\textbf{Ours}} \\
        RAISE 
        & \makecell{\textbf{30.30} \\ {(+7.1\%)}} 
        & \makecell{\textbf{47.98} \\ {(+5.3\%)}} 
        & \makecell{\textbf{51.01} \\ {(+9.8\%)}} \\
        \bottomrule
    \end{tabular}
    \vspace{0.2cm}
    \caption{Evaluation on GPQA with various LLMs with different scales: LLaMA 3.1-8B, GPT-4o mini, and Mistral Small 3.1. Underscore marks the best baseline; bold indicates the best overall. Values in parentheses under RAISE show gains over the top baseline. RAISE consistently shows the best performance across all settings.}
    \label{tab:vertical-llm-eval}
\end{table}

\subsection{Qualitative Evaluation of Retrieved Documents} \label{appendix:qualitative_evaluation}

\begin{figure*}[t!]
\centering
\renewcommand{\arraystretch}{1.3}
\begin{tabular}{p{2.5cm} | p{3.3cm} | p{3.6cm} |p{3.6cm}}
\toprule
\textbf{Subquestion} & \textbf{RAG} & \textbf{RAISE } &\textbf{Explanation} \\
\midrule
What is the product of the reaction of 
2,8-dimethylspiro[4.5]
decan-6-ol with sulfuric acid?
&
Carbylamine reaction ... synthesis of an isocyanide by the reaction of a \sethlcolor{rag}\hl{primary amine, chloroform, and base}. 

&\sethlcolor{raise}\hl{The alkene acts as a nucleophile and attacks the proton, following Markovnikov's rule}. In the second step, an HO molecule bonds to the more substituted carbon... 
&The RAISE-retrieved document explains the acid-catalyzed dehydration mechanism of alcohols, directly aligning with the transformation of 2,8-dimethylspiro[4.5]decan-6-ol to a ketone. 
\\
\midrule
What is the concentration of OH\textsuperscript{--} ions in a solution of 0.3 M Ba(OH)\textsubscript{2}?
&
\sethlcolor{rag}\hl{Normality is an ambiguous measure of the concentration of a solution}. It needs a definition of the equivalence factor...
&
\sethlcolor{raise}\hl{Barium hydroxide is a chemical compound with the formula} Ba(OH)\textsubscript{2}(H\textsubscript{2}O). \hl{Barium hydroxide can be prepared by dissolving BaO in water...} The Ba centers adopt a square anti-prismatic geometry.
&The RAISE-retrieved document clearly identifies barium hydroxide as \(\mathrm{Ba(OH)_2}\) and explains its dissociation behavior in water, directly supporting the calculation of \([\mathrm{OH}^-]\) concentration. 
\\

\bottomrule
\end{tabular}
\caption{Examples where RAISE-retrieved documents provide logically relevant information for scientific reasoning compared to baseline RAG retrieval.}
\label{fig:raise_vs_rag}
\end{figure*}

We qualitatively demonstrate the examples when RAISE retrieves logically relevant documents compared to convential RAG in Figure \ref{fig:raise_vs_rag}. While RAG often retrieves documents that are topically related yet fail to address the reasoning needs of the subquestion, RAISE consistently identifies documents that include essential scientific principles, mechanisms, or equations. For instance, in questions involving chemical reactions, RAISE surfaces materials that explain the specific reactivity or the retarded time calculation, whereas RAG may return vague definitions or unrelated economic concepts. These cases illustrate how RAISE’s retrieval is not only domain-aware but also aligned with the logical demands of solving complex scientific problems.

\begin{figure*}[h!]
    \centering
    \includegraphics[width=\textwidth]{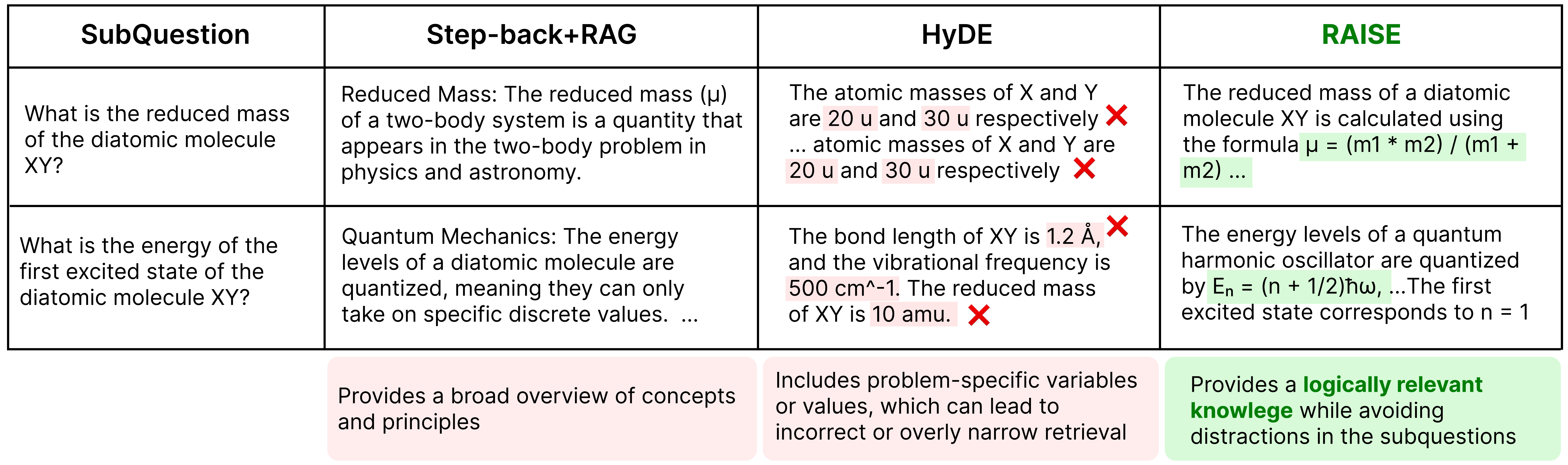}
    \caption{Examples comparing query generation methods (Step-Back+RAG, HyDE, and RAISE) for the same subquestion. Both Step-Back+RAG and HyDE are methods that reformulate the original query to retrieve more relevant documents. These methods are included as baselines in the main comparison table.}
    \label{fig:query_ex}
\end{figure*}

\begin{figure}[ht!]
    \centering
    \includegraphics[width=0.5\columnwidth]{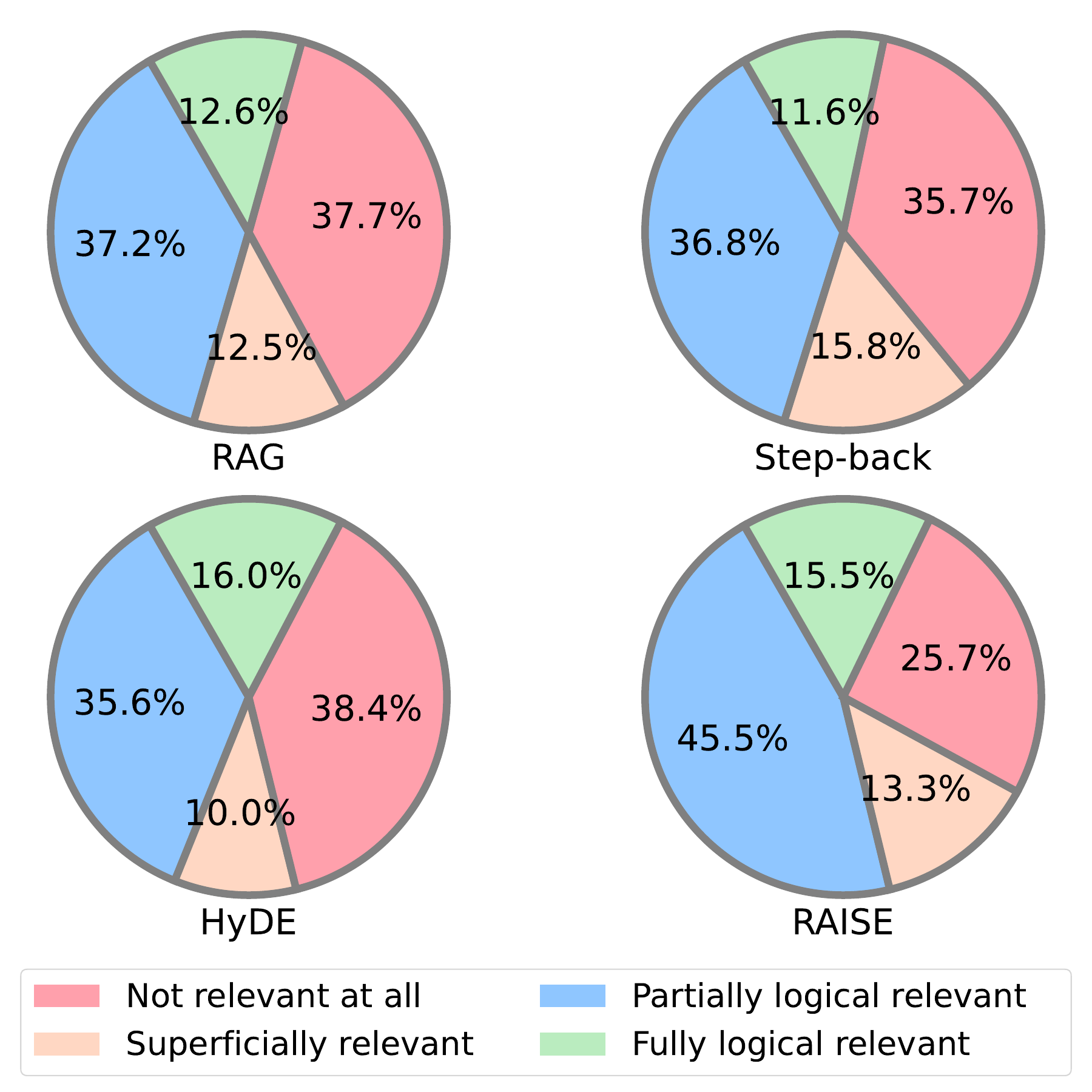}
    \caption{\textbf{Logical Relevancy of Retrieved Documents.} Unlike other baselines, RAISE has higher ratio of documents that are logically relevant and lower ratio of documents that are irrelevant or superficially relevant.}
    \label{fig:logical}
\end{figure}

\begin{figure}[h!]
    \centering
    \includegraphics[width=0.5\columnwidth]{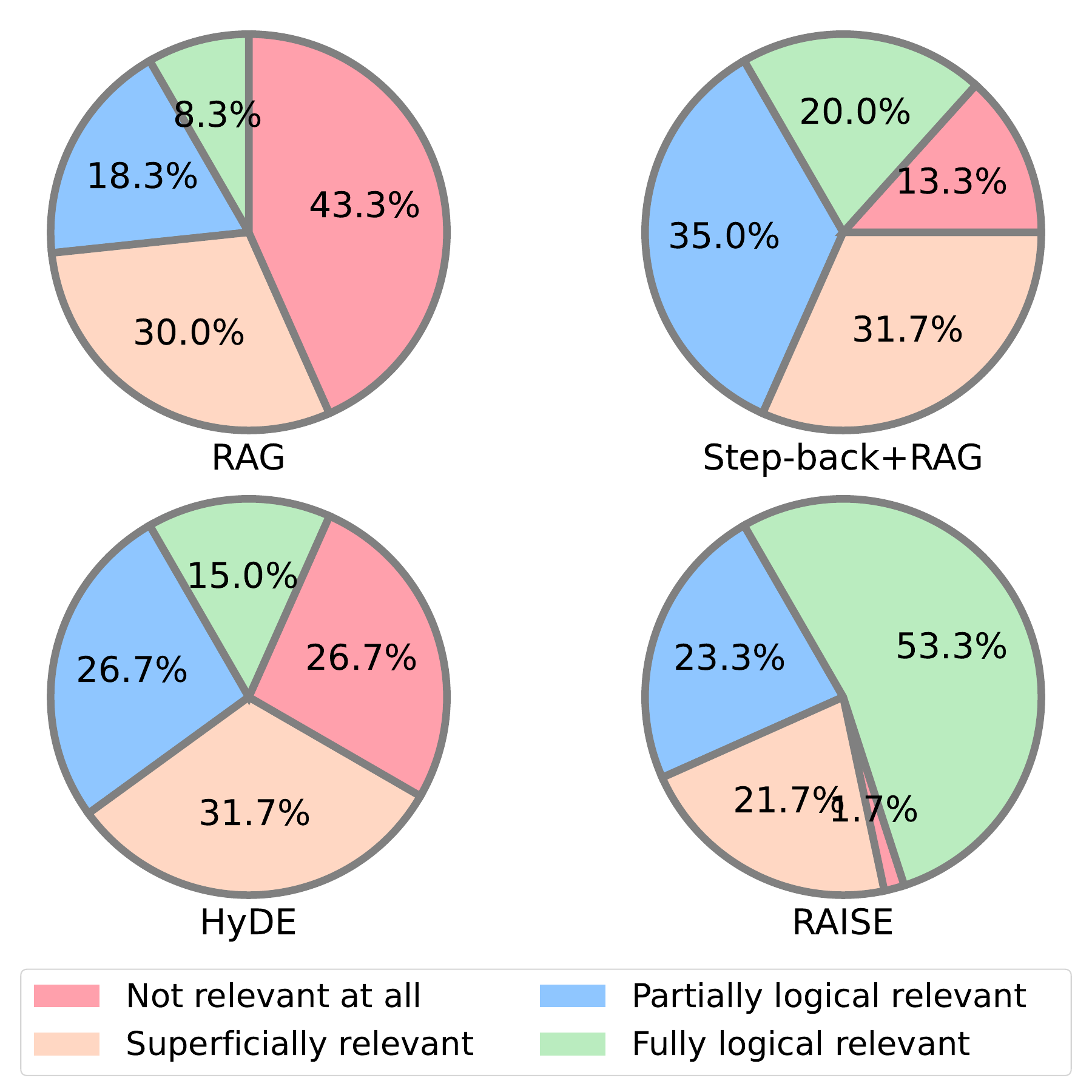}
    \caption{\textbf{Human Evaluation of the Logical Relevance of Retrieved Documents} Aligned with the results from the LLM-as-a-judge evaluation of logical relevancy, RAISE shows a higher proportion of logically relevant documents and a lower proportion of irrelevant or superficially relevant ones.}
    \label{fig:logical_human}
\end{figure}
\subsection{Evaluation of Logical Relevancy of Retrieved Documents} \label{appendix:logical_evaluation}

To further investigate our hypothesis that RAISE retrieves documents that are logically more relevant compared to other baselines, we use LLM-as-a-judge (GPT-4o-mini) to evaluate the logical relevancy of the retrieved documents. Conditioned on the question, subquestion for a specific step, and the retrieved documents, the evaluator model evaluates the logical relevancy among 4 levels of logical relevancy:  (1) \textit{Not relevant at all}, (2) \textit{Superficially relevant} (topically related but logically unhelpful), (3) \textit{Partially logically relevant} (some useful reasoning content), and (4) \textit{Fully logically relevant} (logically sufficient to solve the subquestion).

The results are illustrated in Figure \ref{fig:logical}. Compared to other baselines that also applies RAG, RAISE has the lowest ratio of documents that are irrelevant at all or only superficially relevant (relevant in terms of domain knowledge, but not relevant logically) and highest ratio of documents that are at least partially logically relevant. This indicates that RAISE avoids retrieving documents that may interrupt the reasoning process for scientific reasoning through logical query generation.

Since our domain includes complex, expert-level questions, and LLM-based evaluations may overlook domain-specific reasoning and often rely on surface-level features, we supplemented our analysis with a small-scale human evaluation of 20 subquestion–document pairs. Each pair was assessed by at least three annotators, including Ph.D. students and a faculty member in chemistry, with the method provenance concealed to maintain objectivity.
As also discussed in the LLM-as-a-judge results, the human evaluation indicates that RAISE produces significantly fewer irrelevant documents compared to all other methods, while achieving the highest proportion of logically relevant documents. Although limited in scale due to time and cost constraints, we believe this evaluation provides meaningful human validation of RAISE’s effectiveness and serves as a valuable complement to the LLM-based assessments.

\subsection{Further Analysis on GPQA} \label{appendix:Further analysis on GPQA}
\begin{figure}[h!]
    \centering
    \includegraphics[width=0.8\linewidth]{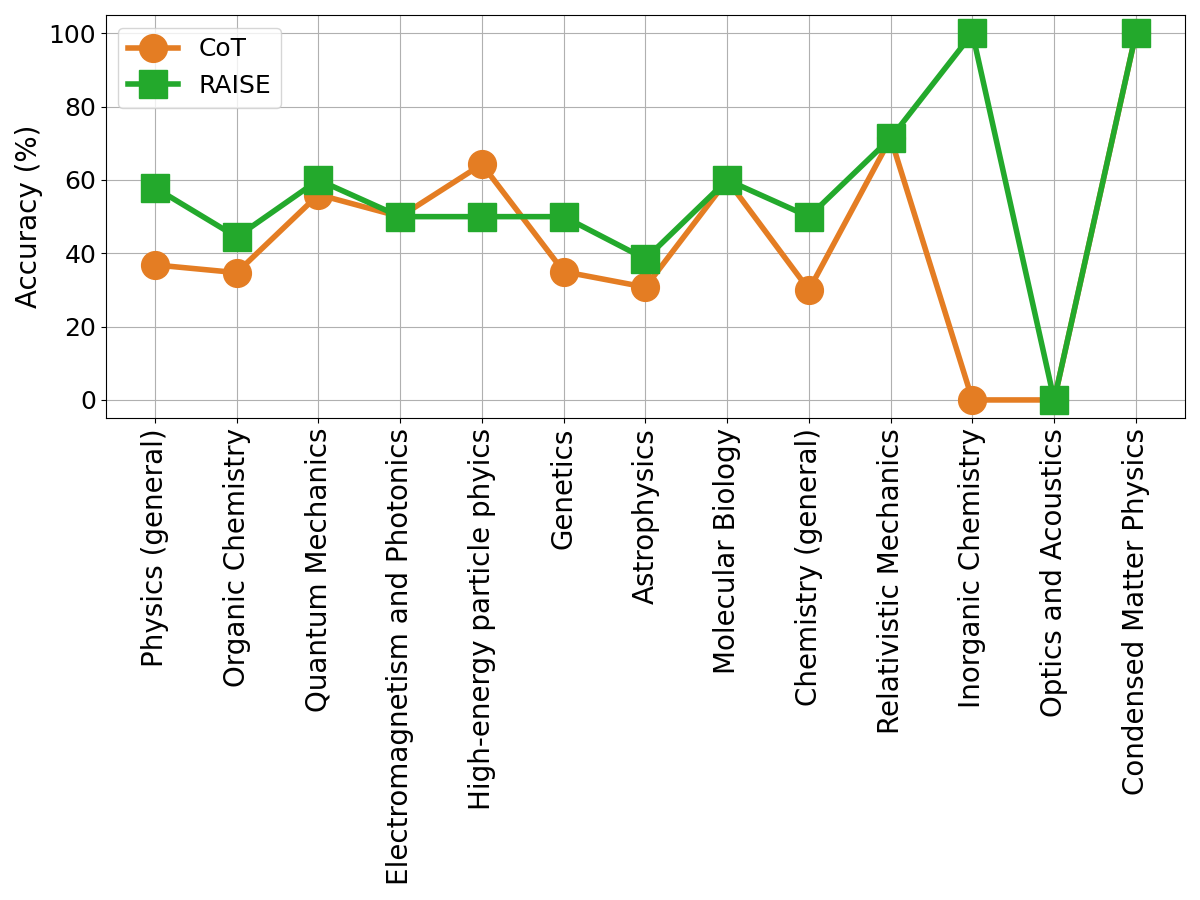}
    \caption{Domain-wise accuracy comparison between CoT and RAISE on the GPQA Diamond subset.}
    \label{fig:gpqa-domain-wise}
\end{figure}

Figure \ref{fig:gpqa-domain-wise} shows the domain-wise accuracy on the GPQA Diamond dataset. We compare the performance of RAISE against Chain-of-Thought (CoT) prompting across all domains. RAISE outperforms or matches CoT in nearly all domains, with only one domain where CoT shows higher accuracy. These results demonstrate RAISE's robustness and its ability to generalize across diverse areas of graduate-level scientific reasoning.

\newpage
\section{Related Works} \label{appendix:related_works}
\paragraph{LLMs for Scientific Reasoning.}
Recent works have shown that LLMs can be applied for challenging scientific reasoning tasks. Unlike other domains, scientific reasoning requires not only step-by-step thinking, but also knowledge of specialized terminology and adaptation to continually evolving knowledge. Due to this challenging nature, many benchmarks have been proposed recently to tackle scientific reasoning with LLMs \citep{gpqa, zhong2025benchmarkingretrievalaugmentedgenerationchemistry, zhang2025physreasoncomprehensivebenchmarkphysicsbased, wang2024scibenchevaluatingcollegelevelscientific, jiang2024visscience}. Many works enhance scientific reasoning capabilities of LLMs through domain-specific training \citep{taylor2022galacticalargelanguagemodel,prabhakar2025omnisciencedomainspecializedllmscientific,zhang2024chemllm}, step-by-step reasoning \citep{rueda2025understandingllmscientificreasoning, gpqa}, or retrieval of external knowledge or tools \citep{ma2024sciagent, zhong2025benchmarkingretrievalaugmentedgenerationchemistry,wellawatte2025chemlit, li2025search}. Unlike previous works, we focus on applying step-by-step document retrieval from in-the-wild corpus without assuming access to well-curated and domain-specific corpus.

\paragraph{Step-wise Reasoning.} 
A growing body of research has shown that decomposing complex problems into structured intermediate steps can enhance the reasoning abilities of LLMs. An influential early approach, Chain-of-Thought prompting \citep{cot}, introduced explicit, sequential reasoning steps, making the model’s thought process more transparent and coherent. This inspired methods such as Plan-and-Solve \citep{pos}, which emphasizes high-level planning before answering, and Step-Back Prompting \citep{step-back}, which encourages abstraction by prompting the model to reflect before solving. Least-to-Most prompting \citep{leasttomost} extends this by breaking down tasks into simpler subproblems, solved in increasing order of difficulty. 

While prior work has focused on prompting strategies that help LLMs better use their internal reasoning capabilities, our work addresses a complementary challenge: enabling LLMs to retrieve and apply information from in-the-wild sources like Wikipedia, particularly during step-wise problem solving. We investigate how external evidence can be integrated at each step to improve reasoning beyond what internal knowledge alone can achieve.

\paragraph{Retrieval Augmented Generation.} 
Retrieval-Augmented Generation (RAG) \citep{rag} was initially proposed to improve LLMs’ factual accuracy and knowledge by retrieving relevant external documents during generation \citep{rag, dpr, passage-retrieval, colbert}.

Recently, RAG has been extended for multi-hop reasoning, performing retrieval iteratively at multiple reasoning steps \citep{hoprag, zhao2024stepwise,step-back}. In parallel, query reformulation and expansion techniques have been developed to enhance retrieval. Instead of using the original question, models generate enriched queries through prompting, such as intermediate answers or summaries. For example, HyDE \citep{hyde} and CSQE \citep{CSQE} demonstrate that carefully crafted queries greatly improve retrieval in complex, multi-step tasks.

Building on this line of work, we redesign query expansion techniques with the specific goal of retrieving documents that contain the key logic or underlying principles required at each step of a step-wise reasoning process. This enables the model to supplement its limited internal knowledge with external sources, leading to more complete problem solving, especially in complex, multi-step tasks.

\clearpage
\section{Prompts} \label{appendix:baselineprompts}

\subsection{Baseline Prompts}

\begin{figure}[h!]
\centering
\renewcommand{\arraystretch}{1.4}
\setlength{\tabcolsep}{10pt}
\begin{tabular}{p{0.95\columnwidth}}
\toprule
You are solving a multiple choice question. Think step by step and show your reasoning clearly. \\

At the end, state your answer in the format: \texttt{"The final answer is (X)"}. \\
Here, X must be the correct letter choice. \\

Question: \textcolor{NavyBlue}{[Problem here]} \\

Answer Choices: \textcolor{NavyBlue}{[Answer choices here]} \\

Solution: \\
\bottomrule
\end{tabular}
\label{fig:cot_prompt}
\caption{Prompt for CoT}
\end{figure}

\begin{figure}[h!]
\centering
\renewcommand{\arraystretch}{1.4}
\setlength{\tabcolsep}{10pt}
\begin{tabular}{p{0.95\columnwidth}}
\toprule
You are an expert at Science. You are given a Science problem. \\
Your task is to extract the Science concepts and principles involved in solving the problem. \\
What are the principles behind this question? \\

End your response with \texttt{"End of generation"} after you answer the instructions. \\

Question: \textcolor{NavyBlue}{[Subquestion here]} \\

Principles Involved: \\
\bottomrule
\end{tabular}
\label{fig:stepback_prompt}
\caption{Prompt for Step-Back Principle Abstraction}
\end{figure}

\begin{figure}[h!]
\centering
\renewcommand{\arraystretch}{1.4}
\setlength{\tabcolsep}{10pt}
\begin{tabular}{p{0.95\columnwidth}}
\toprule
You are an expert at Science. You are given a Science problem and a set of principles involved in solving the problem. \\
Solve the problem step by step by following the principles. \\

At the end, state your answer in the format: \texttt{"The final answer is (X)"}. \\
Here, X must be the correct letter choice. \\

Question: \textcolor{NavyBlue}{[Problem here]} \\

Principles: \textcolor{NavyBlue}{[Principles here]} \\

Answer Choices: \textcolor{NavyBlue}{[Answer choices here]} \\

Solution: \\
\bottomrule
\end{tabular}
\label{fig:stepback_solve_prompt}
\caption{Prompt for Step-Back}
\end{figure}

\begin{figure}[h!]
\centering
\renewcommand{\arraystretch}{1.4}
\setlength{\tabcolsep}{10pt}
\begin{tabular}{p{0.95\columnwidth}}
\toprule
Generate a paragraph that answers the question. \\

End your response with \texttt{"End of generation"} after you answer the instructions. \\

Question: \textcolor{NavyBlue}{[Subquestion here]} \\

Explanation: \\
\bottomrule
\end{tabular}
\label{fig:hyde_query_generation_prompt}
\caption{Prompt for HyDE Query Generation}
\end{figure}

\clearpage

\subsection{RAISE Prompts}

\begin{figure}[h!]
\centering
\renewcommand{\arraystretch}{1.4}
\setlength{\tabcolsep}{10pt}
\begin{tabular}{p{0.95\columnwidth}}
\toprule
You are given a multiple-choice question. \\
Break this problem into essential subquestions that directly help solve the original problem. \\
Each subquestion MUST also include its search query. \\
Each search query should reflect scientific or mathematical knowledge needed to answer the subquestion. \\\\

STRICT FORMAT REQUIREMENTS: \\
1. For each subquestion, you MUST provide exactly two parts in this order: \\
- The subquestion \\
- A search query for that subquestion \\

2. Use EXACTLY this format for each subquestion: \\
Subquestion 1: [your specific subquestion] \\
Search Query for Subquestion 1: [Write a search query someone might realistically use to learn how to answer this subquestion] \\

Question: \textcolor{NavyBlue}{[Problem here]} \\
Answer Choices: \textcolor{NavyBlue}{[Answer choices here]} \\
\bottomrule
\end{tabular}
\label{fig:problem_decomposition_prompt}
\caption{Prompt for Problem Decomposition}
\end{figure}

\begin{figure}[h!]
\centering
\renewcommand{\arraystretch}{1.4}
\setlength{\tabcolsep}{10pt}
\begin{tabular}{p{0.95\columnwidth}}
\toprule
You are given a subquestion and a search query. \\

The search query is a realistic phrase that someone might use to find knowledge or reasoning support to answer the subquestion. \\

Your task is to anticipate what essential scientific or mathematical explanation the search result would contain, and write it concisely (2–3 sentences). \\

Focus only on the core concept or principle that would help answer the subquestion. \\
Avoid restating the subquestion, and do not include unrelated or overly general information. \\

\textbf{Subquestion:} \textcolor{raise_jh}{[Subquestion resulting from Problem Decomposition]} \\
\textbf{Search Query:} \textcolor{raise_jh}{[Search query resulting from Problem Decomposition]} \\

\textbf{Explanation:} \\
\bottomrule
\end{tabular}
\label{fig:logical_query_generation_prompt}
\caption{Prompt for Logical Query Generation}
\end{figure}

\begin{figure}[h!]
\centering
\renewcommand{\arraystretch}{1.4}
\setlength{\tabcolsep}{10pt}
\begin{tabular}{p{0.95\columnwidth}}
\toprule
You are solving a multiple-choice question. The question is decomposed into several subquestions. You will be given: \\
\quad 1. The original multiple-choice question \\
\quad 2. Previous subquestions and their solutions (if any) \\
\quad 3. The current subquestion to solve \\
\quad 4. Documents that are relevant to the current subquestion \\

Your task: \\
- Carefully read the original question, any previous subquestions and their solutions, and the current subquestion. \\
- Use the information from the retrieved documents to solve the current subquestion. \\
- Also use your existing knowledge to solve the current subquestion. \\
- Your solution should be detailed and logically structured. \\

\textbf{Documents:} \textcolor{raise_jh}{[Retrieved document]} \\

\textbf{Question:} \textcolor{NavyBlue}{[Problem here]} \\
\textbf{Answer Choices:} \textcolor{NavyBlue}{[Answer choices here]} \\

\textbf{Previous subquestions and their solutions:} \\
\textcolor{raise_jh}{[Previously generated subquestions and solutions]} \\

\textbf{Current subquestion to solve:} \\
Subquestion \textcolor{raise_jh}{[Step num]}: \textcolor{raise_jh}{[Subquestion]} \\

\textbf{Subquestion \textcolor{raise_jh}{[Step num]} Solution:} \\
\bottomrule
\end{tabular}
\label{fig:solving_subquestions_with_documents_prompt}
\caption{Prompt for Solving Subquestions with Documents}
\end{figure}

\begin{figure}[h!]
\centering
\renewcommand{\arraystretch}{1.4}
\setlength{\tabcolsep}{10pt}
\begin{tabular}{p{0.95\columnwidth}}
\toprule
You are solving a multiple-choice question. The question is decomposed into several subquestions. Each subquestion has already been solved. Your task is to carefully read the original question and the several subquestion solutions, then use them to determine the final answer. Think step by step and then finish your answer with \texttt{"The final answer is (X)"} where X is the correct letter choice. \\
\\
\textbf{Original Question:} \\
Question: \textcolor{NavyBlue}{[Problem here]} \\
Answer Choices: \textcolor{NavyBlue}{[Answer choices here]} \\
\\
\textbf{Subquestions and Solutions:} \\
\textcolor{raise_jh}{[Generated stepwise subproblems and solutions]} \\
\\
\textbf{Final Solution:} \\
\bottomrule
\end{tabular}
\label{fig:final_answer_prompt}
\caption{Prompt for Generating Final Answer}
\end{figure}

\begin{figure}[h!]
\centering
\renewcommand{\arraystretch}{1.4}
\setlength{\tabcolsep}{10pt}
\begin{tabular}{p{0.95\columnwidth}}
\toprule
You are given the following three items: \\

- Original Problem: \textcolor{NavyBlue}{[Problem here]} \\
- Subquestion: \textcolor{NavyBlue}{[Subquestion here]} \\
- Retrieved Document: \textcolor{NavyBlue}{[Document here]} \\

Your task is to evaluate how helpful the retrieved document is for answering the subquestion. \\

Please follow these instructions: \\
- Do not just check if the topic is related. \\
- Instead, check if the document includes information that helps someone reason through and solve the subquestion. \\
- Focus on whether the document supports actual thinking or steps needed to get the answer. \\

Give your final judgment using only one of the following ratings: \\
- \textbf{"No relevance at all"} – does not have any domain similarity \\
- \textbf{"Superficially relevant"} – has domain similarity (only superficially) but does not have any logical relevance to the subquestion. For example, the document might mention the same topic as the subquestion, but it does not provide any information that helps solve the subquestion. \\
- \textbf{"Partially relevant"} – has domain similarity and has some logical relevance to the subquestion. For example, the document might provide some information that helps solve the subquestion, but it does not provide all the logical steps needed. \\
- \textbf{"Fully relevant"} – has domain similarity and has almost all logical relevance to the subquestion. For example, the document provides enough relevant logical steps to solve the subquestion. \\

Then explain your reasoning briefly. \\

\textbf{Output Format:} \\
Helpfulness Rating: \texttt{<one of the 4 options above>} \\
Explanation: \texttt{<your short explanation>} \\
\bottomrule
\end{tabular}
\label{fig:evaluation_with_gpt_prompt}
\caption{Prompt for Evaluation with GPT}
\end{figure}


\end{document}